\documentclass[10pt,twocolumn,letterpaper]{article}

\usepackage{epstopdf}
\usepackage{times}
\usepackage{epsfig}
\usepackage{graphicx}
\usepackage{subfigure}
\usepackage{amsmath}
\usepackage{amssymb}
\usepackage{tabulary}
\usepackage{multirow}

\usepackage{soul}
\usepackage{booktabs}
\begin{document}

\title{Object Activity Scene Description, Construction and Recognition}

\author{Hui Feng$^\dagger$ \\ {\tt\small feng@whut.edu.cn}
\and Shanshan Wang$^{\ddagger}$ \\ {\tt\small eachshan122@163.com}
\and Shuzhi Sam Ge$^{\ddagger}$ \\ {\tt\small samge@nus.edu.sg}
\and $^\dagger$School of Transportation, Wuhan University of Technology
\\$^\ddagger$Department of Electrical and Computer Engineering, National University of Singapore
}

\maketitle

\begin{abstract}
\vspace{-.5em}
Action recognition is a critical task for social robots to meaningfully engage with their environment. 3D human skeleton-based action recognition is an attractive research area in recent years. Although, the existing approaches are good at action recognition, it is a great challenge to recognize a group of actions in an activity scene. To tackle this problem, at first, we partition the scene into several primitive actions (PAs) based upon motion attention mechanism. Then, the primitive actions are described by the trajectory vectors of corresponding joints. After that, motivated by text classification based on word embedding, we employ convolution neural network (CNN) to recognize activity scenes by considering motion of joints as ``word'' of activity. The experimental results on the scenes of human activity dataset show the efficiency of the proposed approach.
\end{abstract}

\vspace{-1em}
\section{Introduction}
\label{intro}
Action recognition is an active topic in computer vision which aims at marking video frames with proper action labels \cite{poppe,Weinland}. It is now widely applied in domains such as human-robot interaction, intelligent perception of social robots, visual surveillance and video retrieval. Action recognition is usually composed of three major steps, i.e., action segmentation, representation and classification \cite{diogo}. Existing works tend to split the actions in the video frame by frame, or by several frames, which results in large amount of features.

We know that action recognition can be seen as an intermediate stage that can provide more complex interpreting systems, such as human behavior analysis and activity scene identification \cite{Magnanimo,Hongsheng}. It is actual that the recognition of the scene of human activity which is composed of a group of actions, is still a highly challenging work currently.

In this paper, we introduce an activity scene description, construction and recognition method based on 3D skeleton sequence to tackle this problem. We at first partitioned the scene of human activity into different primitive actions through the analysis of kinematic of joints. Then, an action descriptor that is able to discriminate the difference of primitive actions was proposed to describe these motions. Finally, motivated by text classification based on word embedding, a convolution neural network (CNN) is exploited to recognize activity scenes by considering motion of joints as ``word'' of activity.

The key contributions of this work can be summarized as follows:
\begin{itemize}
\item We partition the scene of human activity into primitive actions according to the latest research result that the speed information and temporal cues are the two most important factors in tracking moving object. Through conduct the step of scene partition, the features are condensed remarkably.
\item The features extracted from skeleton sequence employed to describe the primitive actions are composed of both spatial and temporal information, which helps to effectively improve the recognition accuracy.
\item By regarding the primitive actions as the ``words'' of human activity, a convolution neural network model which is very efficient in word embedding based topic classification is used to implement scene recognition.
\end{itemize}

This paper is organized as follows. A review of related work is presented in Section \ref{sec_related}. Section \ref{sec_method_main} describes the proposed activity scene description, construction and recognition method, followed by the experimental evaluation in Section \ref{sec_exp}. Finally, we conclude this paper in Section \ref{sec_conclude}.

\section{Related Work}
\label{sec_related}
In this section, we will introduce the state-of-art work involved in action recognition, mainly including the two categories of approaches proposed in the literature in recent years: one was based on 2D video stream, and the other was based upon 3D skeleton joints sequence. In the first category, hand-crafted features, such as histogram of gradient (HOG), histogram of optical flow (HOF) , motion boundary histogram (MBH) and point trajectories, which were popularly used to represent human actions are extracted from raw video stream for action classification \cite{Laptev},\cite{wang}. To overcome limitation of lack semantics and discriminative capacity of these features, deep learning methods were proposed to automatically learn the semantic representation from raw video by using a deep neural network trained from a large amount of labeled dataset \cite{Taylor,Ji,Karpathy,Simonyan,Wang_2}. Recently, to further improve the accuracy of action recognition, neural networks with long-term temporal convolutions with increase temporal extent were proposed \cite{Varol}.

\begin{figure*}
\centering
  \includegraphics[width=14cm]{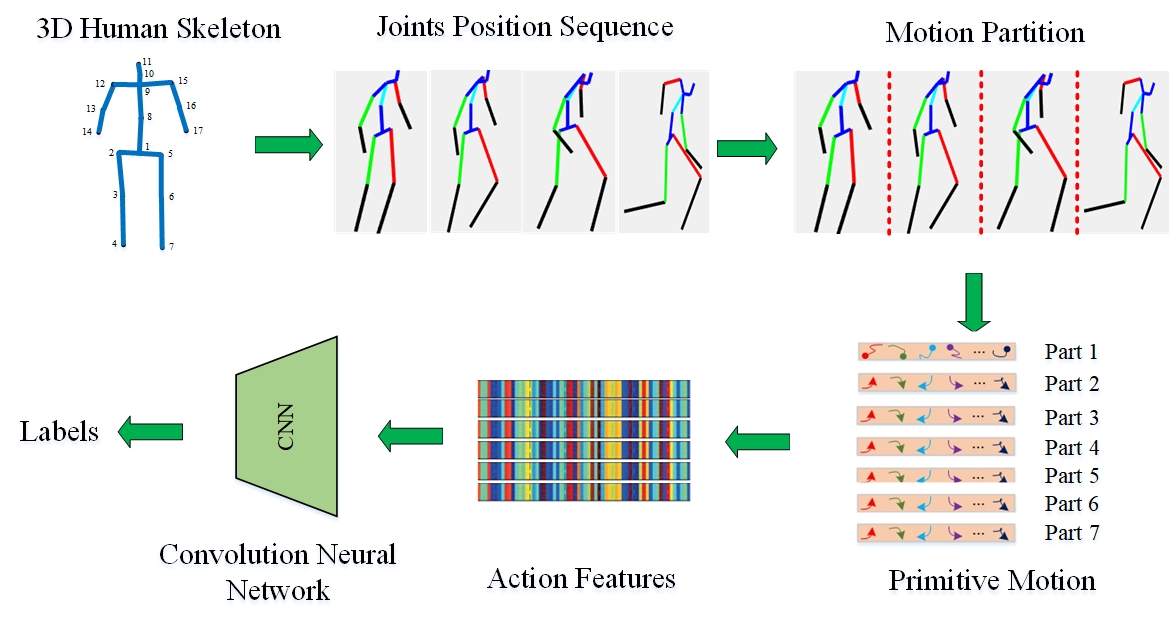}
\caption{The pipeline of the primitive motion partition and description method}
\label{fig:1}       
\end{figure*}

With the rapid development of 3D human pose estimation technologies and the widely usage of depth camera, 3D human poses i.e., skeleton joints, are able to be extracted in real-time \cite{Shotton,Mehta}. Recently, the researchers pay more attention on the methods which are based on 3D skeleton joints. In this category, spatial, temporal or trajectory information is extracted from skeleton joints to represent the features of human action \cite{Xia,Luo,Vemulapalli,Devanne,Slama,Wang_twostream}. The combination of distinct features shows the advantages in building more discriminative features in recent literature \cite{Wang_3,diogo}. Especially, a new framework for action recognition was proposed in \cite{diogo} by means of learning features combination from skeleton sequences. The expressive results had been illustrated in their literature with the average of more than 90.00\% on accuracy. However, several limitations exist in these methods: (i) they just addressed the specifical human actions which are composed of short, simple and well-defined sequence of movements, and (ii) the features extracted from skeleton joints are generated frame by frame which has higher computing complexity.

\section{Object Activity Scene Description, Construction and Recognition}
\label{sec_method_main}
\subsection{Pipeline}
\label{subsec_method_pipeline}
The pipeline of the proposed method is illustrated in Figure \ref{fig:1}. In the first stage, the human skeleton joints are divided into seven parts according to the kinetic analysis of moving joints in action scenes. The detailed analysis is shown in Section \ref{subsec_method_partition}. A recent research proved that, in the process of visually tracking a moving object, humans are generally most sensitive to the speed and time of the moving object, i.e., the speed and time are the two most important factors in object tracking. We call it as motion attention mechanism. Under the guidance of this mechanism, we use the synthetic motion parameter of each group of part as the indicator to partition the primitive motions separately. Then, the representative features are extracted to describe the partitioned motions. Because both spatial and temporal information is considered, these features help to improve the recognition accuracy. In the final stage, a convolution neural network is exploited to recognize the scenes and output the resulting labels. We give the details of this part in Section \ref{subsec_cnn}.

\begin{figure*}[t]
\centering
\subfigure[]{
\includegraphics[width=8cm]{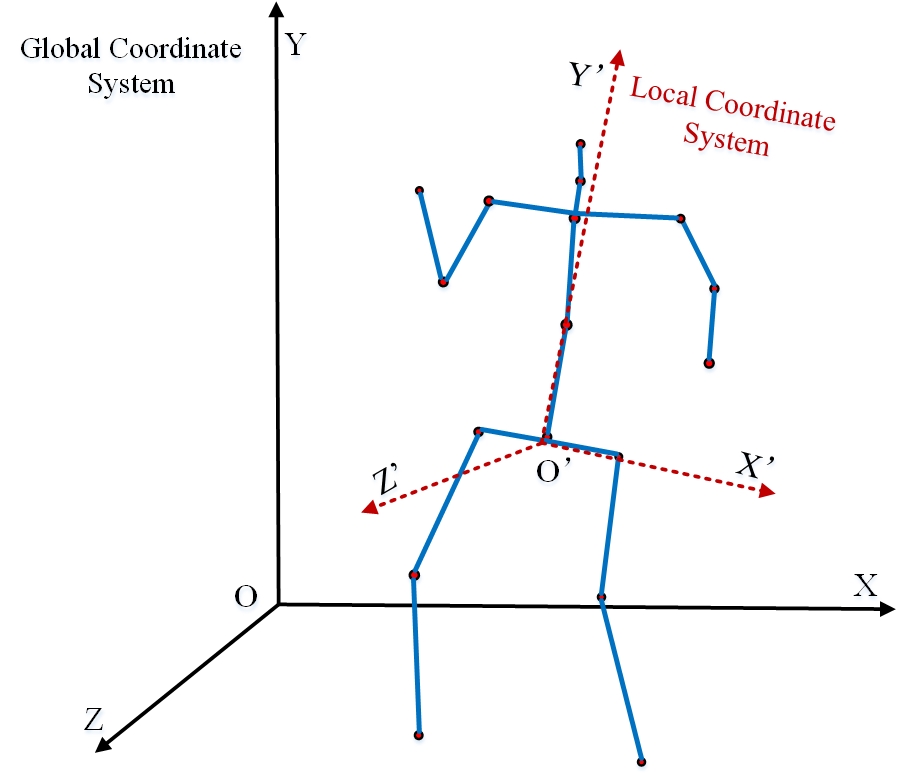}}
\hfil
\subfigure[]{
\includegraphics[width=6cm]{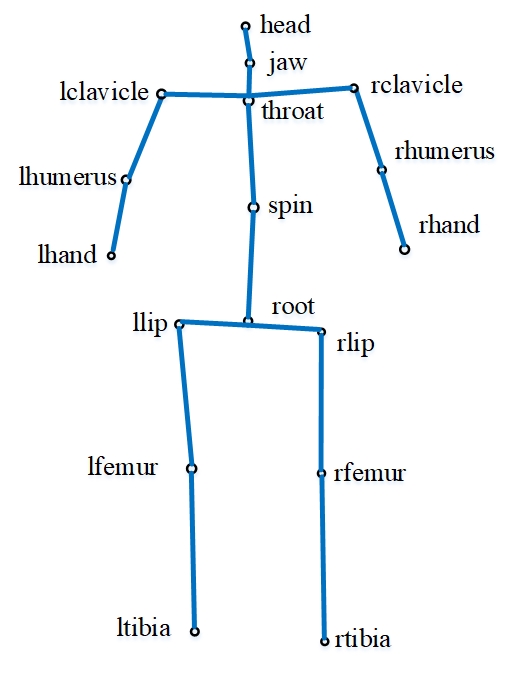}}
\hfil
\caption{(a) Coordinate system. (b) The joints in human skeleton. \label{fig_coor_skel}}
\end{figure*}

\subsection{Object Activity Scene Construction}
\label{subsec_method_partition}
In this section, we introduce the details of how to construct the human actions. Intuitional, a human action consists of several primitive motions which can be regarded as the atomic movements that are used to express specific semantic behavior. Recently, an actionlet ensemble method was proposed to discover discriminative motions by using data mining from several hand-crafted features \cite{Wang_actionlet}. The limitation is that the motions of joints are assumed to be independent. On the contrary, they are coupled to each other and also very complex even if they are described separately.

\subsubsection{Body Parts Division}
\label{subsubsec_division}
We know that human movement includes whole body movement (if we regard human as a particle) and local relative posture (if we take a specific point on the body as a reference). In order to eliminate the disturbance introduced by whole body movement in action recognition, through constructing local coordinate systems, we emphasize the relative motion parameters of the joints under the assumption that majority of action can be recognized by local posture.
\begin{figure*}[t]
\centering
\subfigure[]{
\includegraphics[width=4cm]{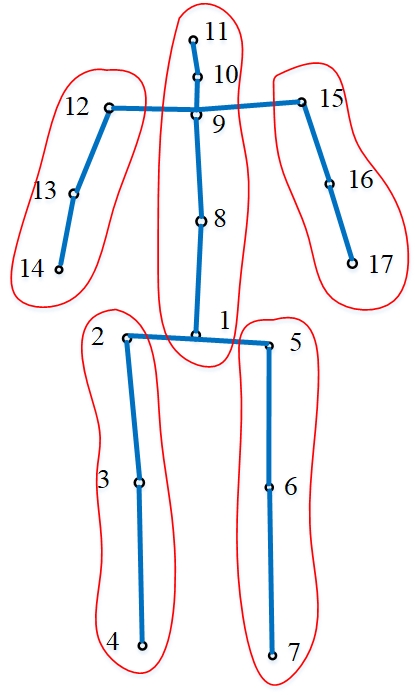}}
\hfil
\subfigure[]{
\includegraphics[width=4cm]{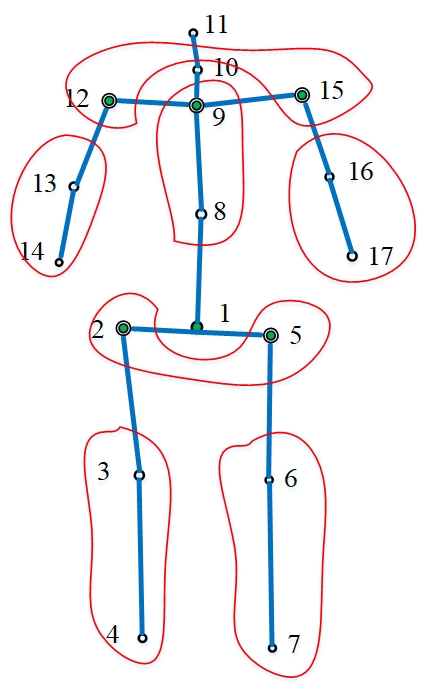}}
\hfil
\caption{(a) Human skeleton joints are divided into five parts in \cite{Du}. (b) Human skeleton joints are divided into seven parts in this paper. \label{fig_partition}}
\end{figure*}

\paragraph{The coordinate system}
To properly describe the motion of human body, two coordinate systems are constructed in this paper, i.e., global coordinate system (GCS) and local coordinate system (LCS), as shown in Figure \ref{fig_coor_skel}(a). GCS is a right-handed coordinate system that places a camera at the origin with the positive $z$-axis extending in the direction in which the camera is pointed \cite{Xia}. Three joints are chosen to construct LCS, i.e., the spin, the left and right hip joints, as illustrated in Figure \ref{fig_coor_skel}(b), because they are more suitable to describe the motion of human body as a whole. The foot drop of the connection of the left and right hip joints and the plain which passes through the spin is set to the origin of LCS (denoted by ``root'' joint), and the positive $x$-axis points to the left hip joint, and $y$-axis starts from the origin to the spin joints. The right-handed rule is also true in LCS.

\paragraph{Construction of human joints division pattern}
From the kinematic perspective, we know that the displacement, the speed and the acceleration are three important indicators of motions. Actually, human actions are extremely difficult to describe, because the motions of human joints have strong arbitrariness and some of them are also highly related. In order to efficiently describe the motions of the joints, a feasible solution is to construct a human joints division pattern, i.e., divide the human skeleton joints into several groups, and then describe them individually.

If we examine each joint separately, the dimension of feature vectors will be large and it will easily increase the computing complexity in the recognition phase. The criterion of dividing human joints relies on two aspects: the joints with the same motion pattern are correctly divided into the same group, while significantly reducing the dimension of the feature vector.

Du et al. \cite{Du} divided all the human skeleton joints into five parts, i.e., two arms, two legs and one trunk, as shown in Figure \ref{fig_partition}(a), but the limitation is that the joints with the different motion pattern are not separate properly which will lead to the degradation of recognition accuracy. For example, we infer that 13 lhumerus joint and 14 lhand joint may have similar pattern, because the forearm is always driven by the upper arm to produce the local motion. In the contrary, 12 lclavicle joint is high related to upper torso which are more suitable to reflect whole body's motion.
\begin{table*}[htbp]
  \centering
  \caption{The motion similarity of selected joints}
    \begin{tabular}{|c||c|c|c||c|c|c|}
    \toprule
    \textbf{Selected Joints} & \multicolumn{1}{c|}{$D_{12,13}$} & \multicolumn{1}{c|}{\textbf{$D_{12,15}$}} & \multicolumn{1}{c||}{$D_{15,16}$} & \multicolumn{1}{c|}{$D_{2,3}$} & \multicolumn{1}{c|}{\textbf{$D_{2,5}$}} & \multicolumn{1}{c|}{$D_{5,6}$} \\
    \midrule
    \textbf{Motion Similarity} & 1.03E+04 & \textbf{4.69E+03} & 8.64E+03 & 9.34E+03 & \textbf{4.11E+03} & 9.31E+03 \\
    \bottomrule
    \end{tabular}%
\label{tab:divPattern}
\end{table*}%

In order to prove our inference, we define a function to measure the similarity of the motions of different joints. At first, we give some definitions. It is assumed that the joint locations $(x, y, z)$ of human skeleton in 3D coordinate can be handled directly. The speed of the joint is defined as follows:
\begin{equation}
\label{equ_1}
v_{j,i}^f = \frac{{s_{j,i}^f - s_{j,i}^{f - 1}}}{{\Delta _t}},i \in (1,2,3)
\end{equation}
where $s_{j,i}^f$ is the location of $j$-th joint in $i$-th coordinate axis and $f$-th frame, and $\Delta _t$ is the time interval between two frames.

Then, the synthetic speed of the joints in three directions is denoted by
\begin{equation}
v_j^f = \sqrt {\sum\limits_{i = 1}^3 {{{(v_{j,i}^f)}^2}} }
\end{equation}
and the corresponding synthetic acceleration of $j$-th joint can be written as
\begin{equation}
a_j^f = \frac{{v_j^f - v_j^{f - 1}}}{{\Delta _t}}
\end{equation}
The Euclidean distance between two joints' synthetic acceleration is employed to measure their motion similarity, and the function is given by
\begin{equation}
\label{equ_corelation}
{{D}_{j,k}} = \textbf{dist}({{\bf{a}}_j},{{\bf{a}}_k})
\end{equation}
where ${\bf{a}}_j, {\bf{a}}_k$ are the sequence of synthetic acceleration for joint $j$ and $k$, respectively.

We calculate the motion similarity of several selected joints using the function (4), and the results are shown in Table \ref{tab:divPattern}. From Table \ref{tab:divPattern}, we can see that $D_{12,15}$ is much smaller than $D_{12,13}$ and $D_{15,16}$, and the same results can be observed in hip joints (2 and 5) and leg joints (3 and 6).

\begin{figure*}[htb]
\centering
  \includegraphics[width=18cm]{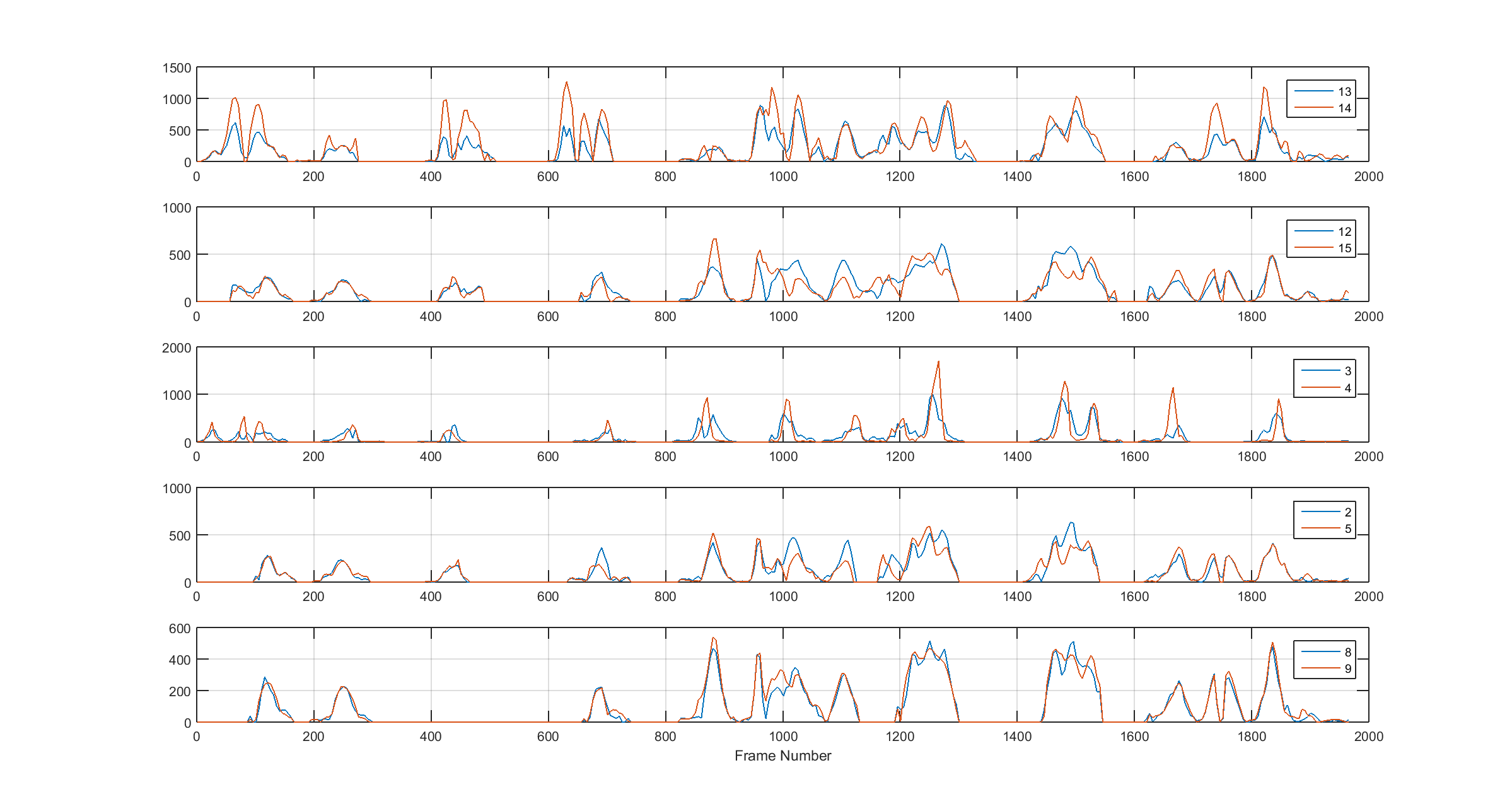}
\caption{The synthetic speed for human skeleton joints}
\label{fig_synthspeed}       
\end{figure*}
To show this similarity more intuitively, we draw the curves of the synthetic speed for part of human skeleton joints, which is illustrated in Figure \ref{fig_synthspeed}. From Figure \ref{fig_synthspeed}, we can see that the motions of 13 lhumerus joint and 14 lhand joint have higher similarity than 12 lclavicle joint on time sequence, but 15 rclavicle joint is higher related to 12 lclavicle joint. The similar pattern can be noticed in the motion of other joints, which uncovers the fact that the function defined in Equ. (\ref{equ_corelation}) represents the inherent characteristic of motion patterns of these joints.
\begin{table*}[htbp]
  \centering
  \caption{Human skeleton joints are divided into seven groups}
  \label{table_division}
    \begin{tabular}{|c|c|c|c|}
    \toprule
  \multicolumn{1}{|c|}{\multirow{2}[4]{*}{\textbf{Sequence Number}}} & \multirow{2}[4]{*}{\textbf{Parts}} & \multicolumn{2}{c|}{\textbf{Joints}} \\
\cmidrule{3-4}          & \multicolumn{1}{c|}{} & \textbf{Pivot joint} & \textbf{End joints} \\
    \midrule
    1     & Lower torso part & root  & lhip, rhip \\
    \midrule
    2     & Spine & root  & spine, throat \\
    \midrule
    3     & Upper torso part & throat & lclavicle, rclavicle \\
    \midrule
    4     & Left arm & lclavicle & lhumerus, lhand \\
    \midrule
    5     & Right arm & rclavicle & rhumerus, rhand \\
    \midrule
    6     & Left leg & lhip  & lfemur, ltibia \\
    \midrule
    7     & Right leg & rhip  & rfemur, rtibia \\
    \bottomrule
    \end{tabular}%
\end{table*}%
According to the above observation, seven parts are divided in our paper which are listed in Table \ref{table_division} and shown in Figure \ref{fig_partition}(b), and their corresponding pivot joints (or rotation center) are also illustrated (green solid points). Torso joints are usually used to reflect the global motion of human, such as move forward, backward, left, right, and jump up and down. In the contrary, limbs joints are more suitable to represent local motions. In this paper, the head motion is ignored.

\subsubsection{The primitive motion partition}
\label{subsubsec_motion}
It is a challenge to describe human's action, because the velocity and position of human joints have strong dependency. Supposing we consider recognition problem as a function approximation from mathematic perspective (which is reasonable when we use neural network to tackle this problem), motivated by defining primitive functions in the parameters neural networks in early literature \cite{Ge_1,Ge_2}, we attempt to partition the scene of human activity into primitive actions.

Through the observation of human action from frame level, we find that each human part (as listed in Table \ref{table_division}) usually has a standstill between two different motions. Two successive standstills can be used to represent the start and the end of a primitive motion, therefore, in this paper, we consider to use the standstill as the indicator of the interval of two primitive motions. The end joints in seven human parts are chosen to partition the motions, because these joints in each part are high related and have the similar motion patterns as explained in Section \ref{subsubsec_division}.

In this paper, the synthetic speed of the joints is employed to partition the actions, which has been defined in Equ. (\ref{equ_corelation}). We notice that the joint positions should be preprocessed to eliminate the negative effect of wild values, meanwhile, insignificant motions should be also filtered to simplify the action recognition. The preprocessed synthetic speed is
\begin{eqnarray}
\bar v_j^f = \left\{ {\begin{array}{*{20}{c}}
{v_j^f,{\rm{  }}{\qquad if \quad}{\rm{ }}v_j^f \ge {v_\tau }}\\
{0,{\rm{  }}{\qquad otherwise \quad}}
\end{array}} \right.
\end{eqnarray}
where ${v_\tau }$ is the threshold which is used to suppress negligible motions.

The recent research of the psychophysicists from MIT revealed that, in the process of visually tracking a moving object, humans rely on both speed information and temporal information \cite{Chang}. That means humans are generally more sensitive to the speed and time of the moving object, which can be considered as \emph{motion attention mechanism}. That means when partitioning primitive actions, the joints with higher moving speed and longer displacement (represented as the product of speed and time) will be paid much more attention. Based on the above consideration, in the time sequence of moving joints, the intervals at which the synthetic velocity of joints has a larger value and a greater width will be considered as the primitive motions.

\begin{figure*}
\centering
  \includegraphics[width=15cm]{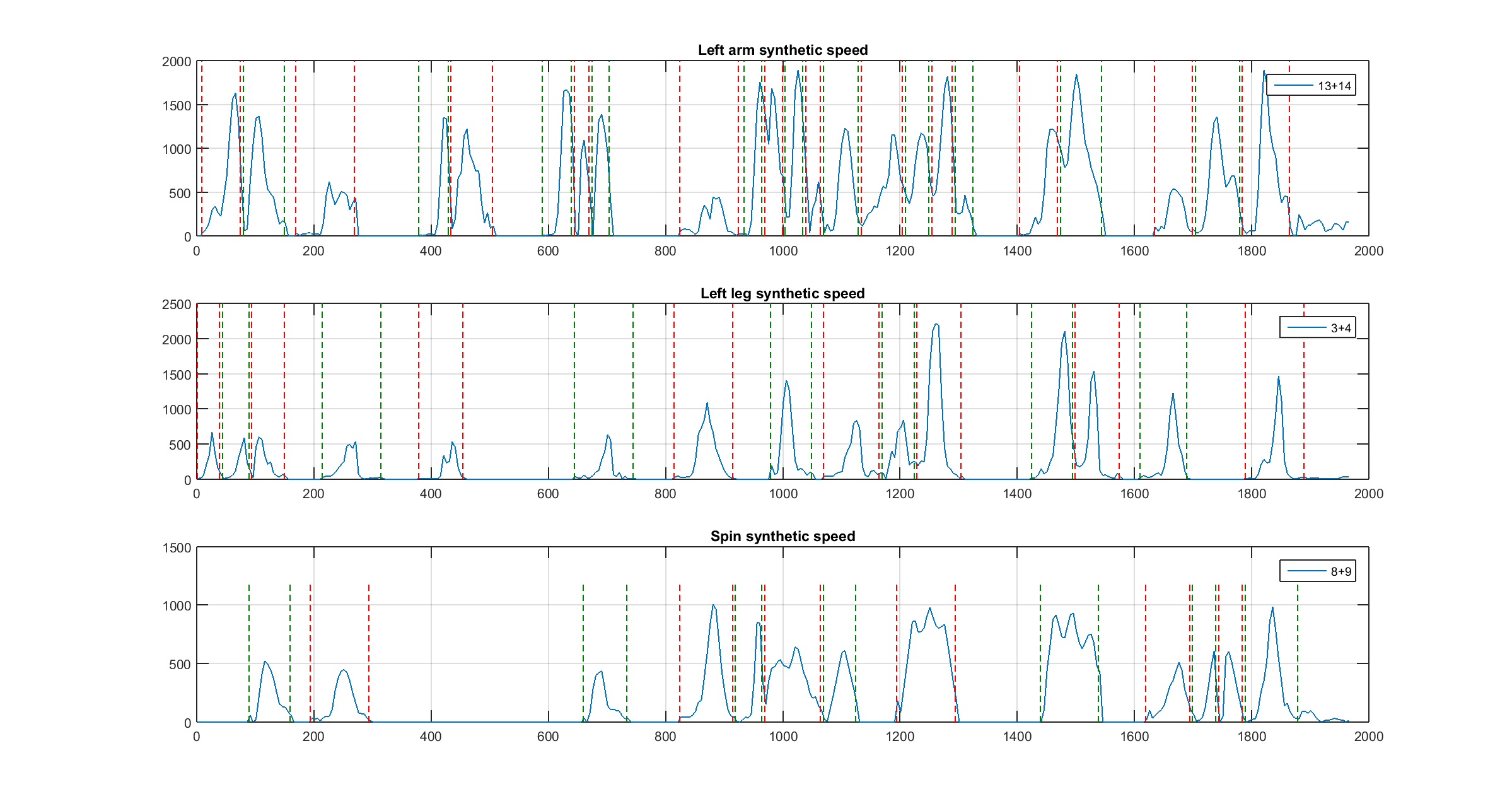}
\caption{The partition results for the primitive motions}
\label{fig_pms}       
\end{figure*}

For the reason of each human part has two joints, the intervals of the two joints are combined, which are denoted as follows:
\begin{equation}
{M_{p,q}} = \mathop  \cup \limits_n ({S_n}|\bar v_j^f \ne 0),n = \{ 1,2\}
\end{equation}
where $n$ is the number of joints in the part, and $M_{p,q}$ is the $q$-th combined interval of the $p$-th human part, and $S_n$ is the interval which had been selected as representing primitive motion. The motions in each combined interval are partitioned as a primitive motion for this part. The number of primitive motions in the scenes can be regarded as a super-parameter. The partition results are shown in Figure \ref{fig_pms}, where only three sets of end joints are drawn as an example. The adjacent primitive motions are distinguished by different colored dotted lines.

\subsection{Primitive motion descriptor}
\label{subsec_descrip}
After partition the primitive motion of these joints, it is extremely important to describe them for the following recognition stage. In this section, we present a primitive action descriptor that consists of two parts: global motion descriptor and local motion descriptor.

\subsubsection{Global motion descriptor}
\label{subsubsec_global}
\begin{figure*}[t]
\centering
\subfigure[]{
\includegraphics[width=7cm]{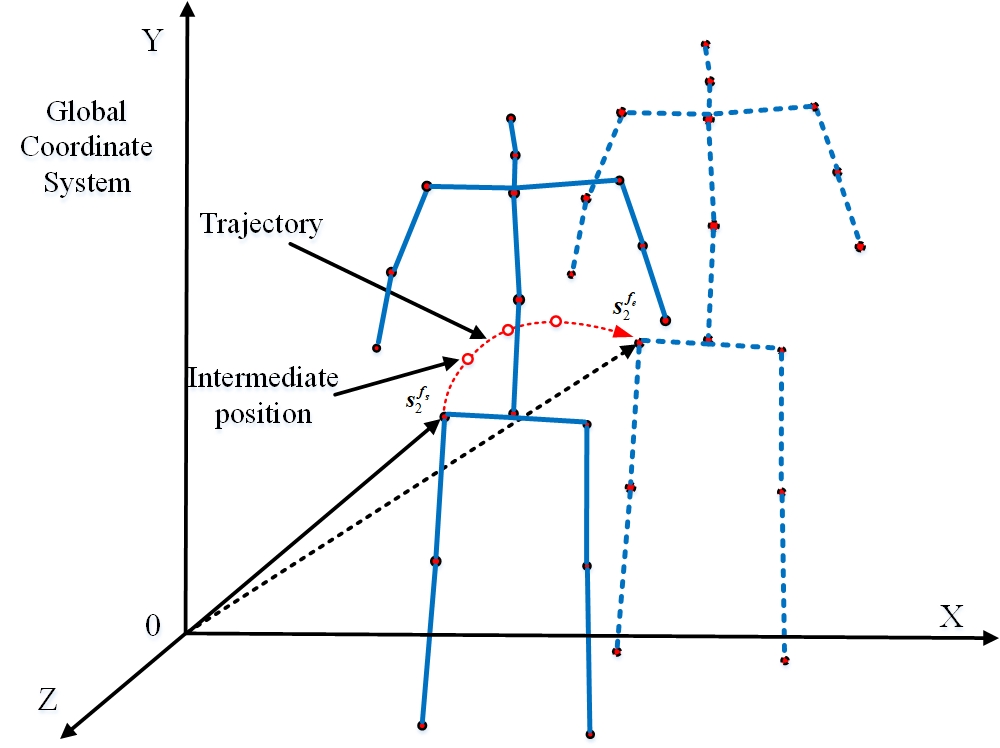}}
\hfil
\subfigure[]{
\includegraphics[width=4cm]{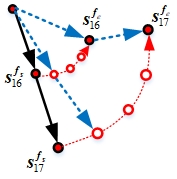}}
\hfil
\caption{(a) Motion description of human torso. (b) Motion description of human limbs. \label{fig_global_limbs}}
\end{figure*}

Global motion is described by three joints, i.e., root, lhip and rhip joints, in lower torso part. To avoid redundant, we just consider lhip and rhip joints. The global features are extracted from the trajectory of these two relevant joints. The joint trajectory is a sequence of points in 3D space that describes the motion path of the specified joint, and it can be represent by a function
\begin{equation}
\label{equ_global}
T_j^q = f(b_j^q,m_j^q,e_j^q)
\end{equation}
where $T_j^q$ is the trajectory of the $j$-th joint in the $q$-th primitive motion, and $b_j^q$ and $e_j^q$ are the start and end displacement vectors related to the origin of GCS which are denoted as
\begin{equation}
b_j^q = s_j^{{f_s}} - {\bf{0}},e_j^q = s_j^{{f_e}} - {\bf{0}}
\end{equation}
respectively, and $m_j^q \in {R^3}$ is the intermediate points in the trajectory, as shown in Figure \ref{fig_global_limbs}(a). In this paper, we choose five intermediate points with uniform distribution from the trajectory. For simplicity, we drop the superscript in the following section.

\subsubsection{Local motion descriptor}
\label{subsubsec_local}
The description of local motion has some difference with global motion. For one thing, the different parts of human body should be described separately. That means the local motion features are composed of totally of seven parts' features. For another, we are just interested in the relative motions of human parts in local environment. The relative trajectory can be denoted by
\begin{equation}
\label{equ_local}
r{T_j} = f(r{b_j},r{m_j},r{e_j})
\end{equation}
The difference between Equ. (\ref{equ_local}) and (\ref{equ_global}) is that the parameters of $j$-th joint in (\ref{equ_local}) are all related to the corresponding pivot joint. We take right arm as an example in Figure \ref{fig_global_limbs}(b), and its features are composed of rhumerus's trajectory vector $r{T_{16}}$ related to rclavicle, and rhand's trajectory vector $r{T_{17}}$ related to rhumerus. The same strategy is used in the selection of intermediate points as in global motion description.

\paragraph{Extension of Local Features}
When we obtain the features of each part according to Equ. (\ref{equ_local}), the motion of end joints in this part is clarified. However, in a natural sense, different parts of human body are bonded rather than isolated when they are combined to represent a specific activity. In this paper, we construct this relationship between the joints in different parts through extending the local features of each part not only in relation to the pivot joint but also in relation to the joints in other parts. After that, the local features of the primitive motions for each part are concatenated, and the features for global and local motions are overlaid according to chronological order.

From the features that used to describe the primitive motions, we can find that both spatial and temporal information are contained. The temporal relationship is reflected in the primitive motion vectors of each part which are arranged by chronological order, while extended local features maintain the spatial bonding between different parts.

\subsubsection{Feature Normalization}
\label{subsubsec_norm}
It is note that the local features extracted from human motion depicted in Section \ref{subsubsec_local} are wholly relative displacement vectors. That means they have latent drawback when comes to the difference of human body. For example, the features would have salient distinction while they are extracted from the same action but conducted by an adult and a kid, because the results are affected by the length of their bodies. To eliminate the impact of difference of human body, the normalization step is followed by motion description. For each displacement vector, the normalized feature is generated by dividing the norm of the corresponding vector.

\subsection{Scene Recognition based on Convolution Neural Network}
\label{subsec_cnn}
We know that before language was invented, action was the most important manner for humans to communicate with others. Human beings deliver specific semantics through a series of actions that had spatial and temporal relationships. It is reasonable to regard these actions as body language comparing to words in natural language. Correspondingly, a group of actions, i.e., activity scene, that is used to deliver sematic can be considered as a ``paragraph''. The task of recognizing the scenes of human activity from the video is similar to classifying a paragraph of words into different topics, i.e., text classification.

\begin{figure*}
\centering
  \includegraphics[width=16cm]{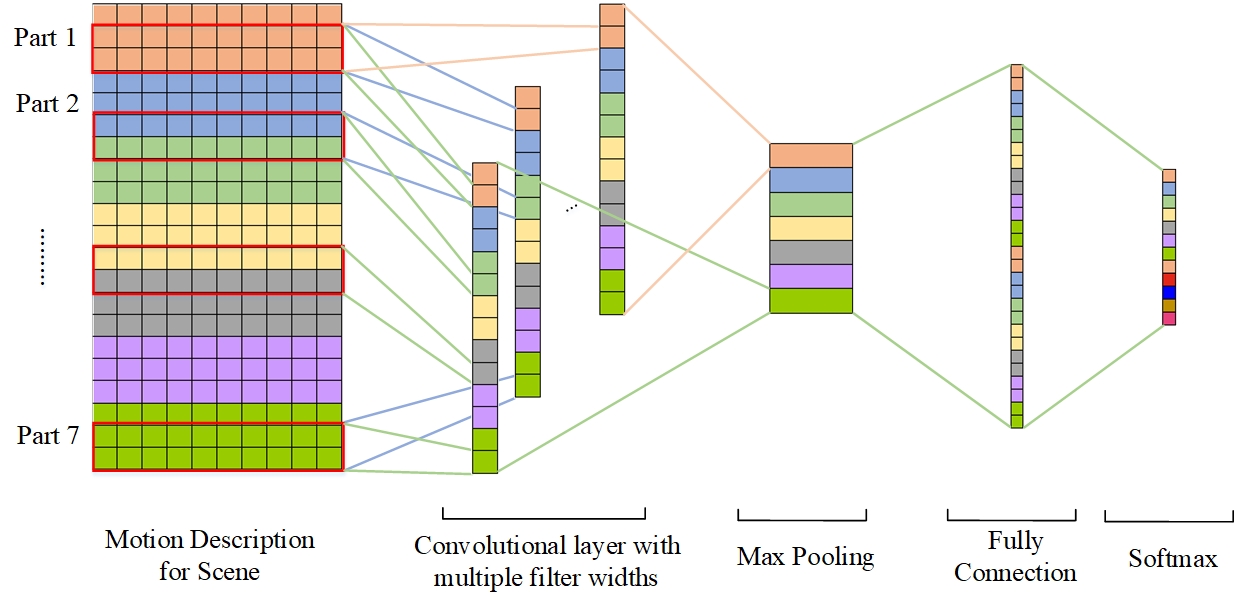}
\caption{The CNN model for scene recognition}
\label{fig_cnn}       
\end{figure*}

Word embedding combined with CNN is an effitive approach popularly used by authors in the field of text classification. Motivated by text classification, we employ CNN to recognize activity scenes by considering actions of joint as ``words'' of activity. CNN has been employed in human action recognition in recent articles because of its promising performance in classification, however, the majority of existing approaches were conducted on raw image sequence. In this paper, our proposed mothed is based on 3D skeleton joints and the primitive action description and representation introduced in the above subsections. The CNN model we constructed is shown in Figure \ref{fig_cnn}. As can be seen from Figure \ref{fig_cnn}, four layers deep neural network is contained in the model with the order of convolutional layer, max pooling layer, fully connection layer and softmax layer. The convolutional layer with multiple filter widths has the capability of learning spatial and temporal relationships of the primitive actions from the features we extracted.

\section{Experimental results}
\label{sec_exp}
It is notice that there are plenty of datasets available for 3D skeleton or depth map based action recognition. Existing algorithms usually evaluate their performance on three popular datasets: MSR-Action3D \cite{Li_msr}, UTKinect-Action3D \cite{Xia} and Florence 3D Actions \cite{Seidenari}, and their average accuracy is up to 98\% \cite{Oreifej,Yang,Du,diogo}. However, these databases are relatively simple because the actions only contain the movement of a single component (left arm, right arm, for example) or a combination of a small number of components. In our experiments, we consider a more challenging activity scene recognition task.

\subsection{H36m Dataset}
\label{subsec_h36m}
The H36m dataset contains 3.6 million 3D human skeleton and corresponding frames collected by 4 digital cameras, 1 time-of-flight sensor and 10 motion cameras, with 11 professional actors (6 male and 5 female) and 17 scenes. We use publicly opened subset with 422,055 frames which are downsampled from original videos (50 fps) (in order to reduce the correlation of consecutive frames). The subset contains 15 scenes associated with 7 subjects whose ground-truth 3D skeleton are provided. The 15 activity scenes include: discussion, direction, eating, greeting, posing, purchases, sitting, sitting down, smoking, waiting, walking, walk together, walk dog, taking photo, and talking on the phone. Each scene contains a series of actions for expressing specific semantic human activity that is closer to the natural interaction scenarios. This dataset is much more challenging because: (1) the lengths of sequences are quite long on average and vary greatly (from 990 to 6340 frames); (2) the diversity within the same class is large, e.g., for ``posing'', different people pose according to their own understanding; (3) the dataset contains confusing actions such as walking, walk together and walk dog, as well as sitting and sitting down.

\subsection{Dataset Augmentation}
\label{subsec_aug}
Although deep neural networks have made great achievement in wide aspect of intelligent tasks, such as text and image classification, natural language processing and action recognition, one of the main challenges of using neural networks in 3D action recognition is that the number of available training samples is relatively small, often leading to overfitting. Intuitionally, seeking the way to augment the dataset has higher priority to eliminate the drawback of overfitting. However, it is difficult to extend the existing datasets because of the bias would inevitably introduced during collecting, processing and validating of new data samples.

We assume that the same semantics can be conveyed when people perform the same action using left limbs and right counterparts. This is always true in real human-human interactions and human-robot interactions. For example, some people like to drink water with their left arms, but others prefer to using their right arms. Resorting to the spatial properties of 3D skeleton joints, we are able to construct the mirror model of human actions. Based on the local coordinate system established in Section \ref{sec_method_main}, we use the $yoz$ plane as the symmetry plane to flip the coordinates of the joints of the human body to obtain the new coordinates of each joint, thereby doubling the original sample size.

\subsection{Parameter Setting}
\label{subsec_param}
Through augmenting the original H360m dataset, the total of 1680 scene samples can be obtained. We split them into three parts: 5 subjects (1200 scenes) as training set, and the rest 2 subjects (480 scenes) as validation set and testing set, respectively.

\begin{figure}
\centering
  \includegraphics[width=7cm]{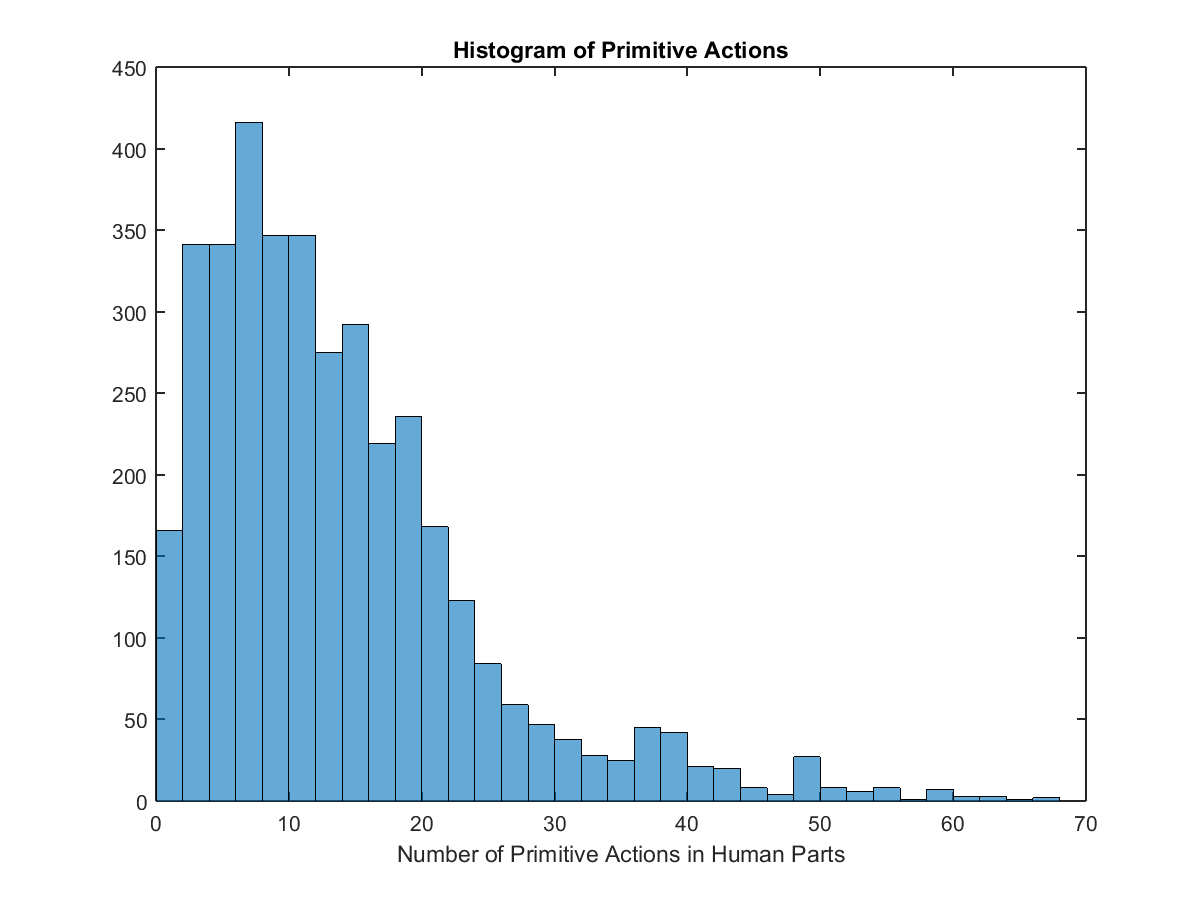}
\caption{The histogram for primitive actions}
\label{fig_hist_pano}       
\end{figure}

To determine the total number of the primitive actions aforementioned in Section \ref{subsubsec_local}, we statistic the primitive actions for human parts extracted from the dataset, as shown Figure \ref{fig_hist_pano}. We can see that the number of PAs lays in interval between 0 to 40, therefore, the maximum of PAs is set to 30 to balance the trade-off in the efficiency and compactness of features. The total dimension of the features for each scene is $240 \times 510$.

For our method, to reveal the sensitiveness of performance on parameters, we illustrate three groups of experimental results under different number of filters in convolutional layer as well as different learning rate. In the first experiment, we fix the learning rate to $10^{-4}$ and set filters number as 768. Then, we fix the number of filters to 1024, and conduct the rest two experiments with the learning rates of $10^{-3}$ and $10^{-4}$, respectively. The number of neurons for fully connection layer is 256 and the implementation is based upon Tensorflow.

As we know, many distinguished action recognition approaches have been proposed in recent years. It is difficult to evaluate all their performances in recognizing activity scenes in H36m dataset. In this paper, only five represented methods are chosen to make the comparison, as shown in Table \ref{tab:perform}.

The average frames of H36m dataset is about 500, so the fixed feature length $T$ for the spatial-temporal RNN \cite{Wang_twostream} to 500. The other parameters of these methods are coincide with the original articles. Specially, for six layer RNN, a deep LSTM model is used and the dimension of hidden neurons is set to 256. All the simulations are run on Intel Core i7-4790 CPU with 32G memory and Nvidia 1080 Ti GPU.

\begin{figure}
\centering
  \includegraphics[width=8cm,bb=0 0 598 495]{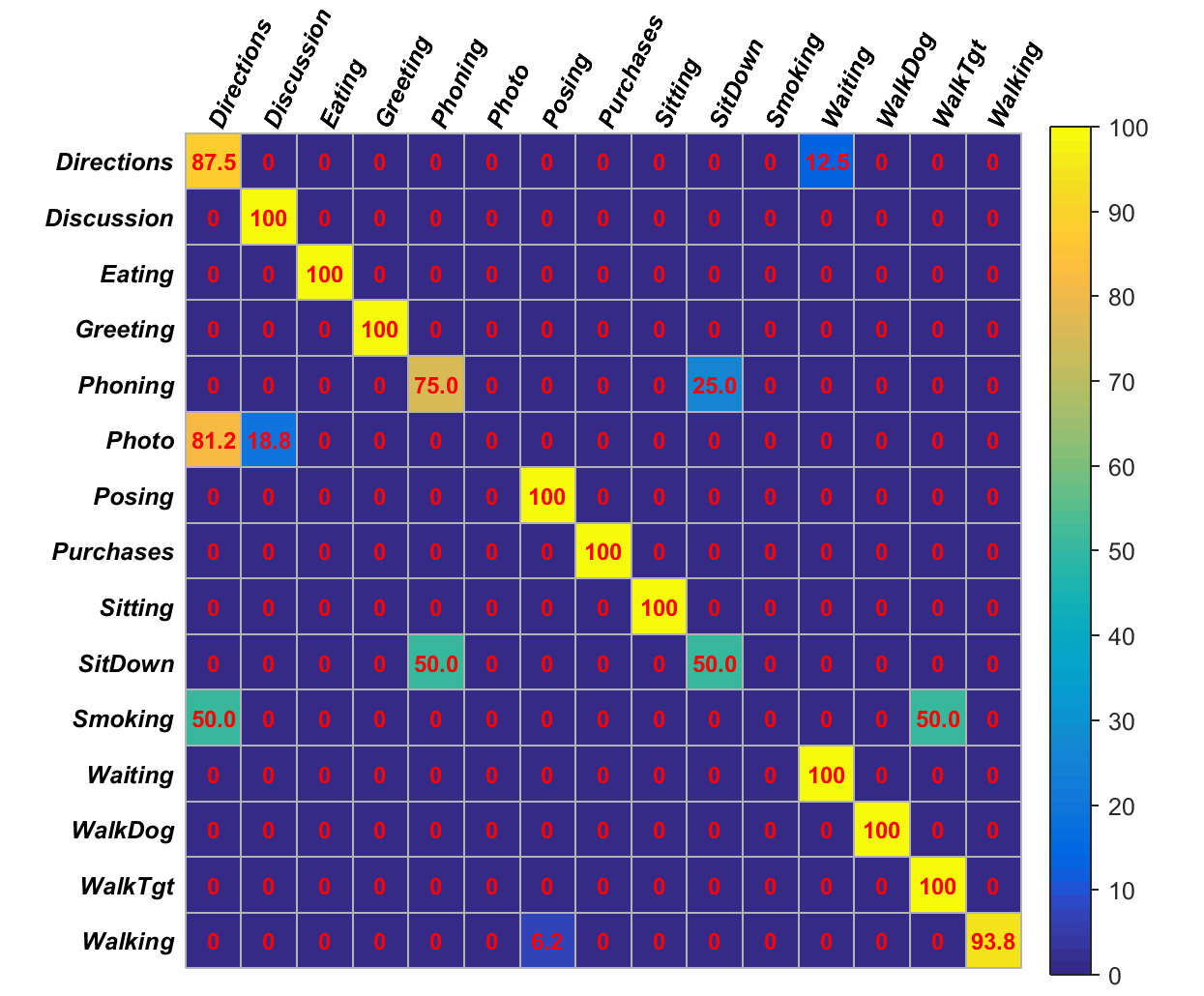}
\caption{The confusion matrix for scene recognition on H36m dataset from the proposed method}
\label{fig_confuse}       
\end{figure}

\subsection{Results Comparison and Analysis}
\label{subsec_results}
The experimental results are shown in Table \ref{tab:perform}. From Table \ref{tab:perform}, we can see that the accuracy of the proposed method is 6.64\% higher than the best result based on vector of locally aggregated descriptors (VLAD) with metric learning \cite{diogo}, and our method also outperforms LSTM based spatial-temporal RNN (7.72\%) and six layers RNN (33.9\%). Because of the inclusion of trust gate in the LSTM, the spatial-temporal RNN has much better improvement (26.18\%) than traditional six layers RNN.

It can be also observed that the performance of the proposed method is more sensitive on the learning rate than the number of filters in the model. The confusion matrix of our method is shown in Figure \ref{fig_confuse}. Several scenes are recognized without any mistake, and most of accuracy is high. It is notice that the classification of two scenes (photo and smoking) is completely wrong, because the actors added too many extra actions that were similar to the other scenes. Even humans do not easy to recognize them from the original videos. This is the reason why H36m dataset is more challenge than the others.

\begin{table}
\centering
\caption{Experimental results on the H36m dataset}\
\label{tab:perform}       
\begin{tabular}{ll}
\hline\noalign{\smallskip}
\textbf{Method} & \textbf{Precision}  \\
\noalign{\smallskip}\hline\noalign{\smallskip}
Lie Group + SVM \cite{Vemulapalli}   & 49.80\% \\
VLAD + Metric Learning \cite{diogo}  & 73.77\% \\
\hline
Six Layers RNN   & 46.52\%   \\
Spatial-temporal RNN \cite{Wang_twostream}   & 72.70\%    \\
\hline
Proposed CNN FN=$768$      & 75.00\%       \\
Proposed CNN LR=$10^{-3}$      & 76.76\%       \\
\textbf{Proposed CNN LR=$10^{-4}$}      & \textbf{80.42\%}       \\
\noalign{\smallskip}\hline
\end{tabular}
\end{table}	

\begin{table}[htbp]
  \centering
  \caption{The computation time}
  \label{tab:time}%
    \begin{tabular}{|c|c|c|c|}
    \toprule
    \multirow{2}[2]{*}{\textbf{Methods}} & \multicolumn{1}{c|}{\textbf{Feature}} & \multicolumn{1}{c|}{\textbf{Training}} & \multicolumn{1}{c|}{\textbf{Testing}} \\
    \multicolumn{1}{|c|}{} & \multicolumn{1}{c|}{\textbf{Extraction}} & \multicolumn{1}{c|}{\textbf{/Epoch}} & \multicolumn{1}{c|}{\textbf{/Sample}} \\
    \midrule
    Lie Group & \multirow{2}[2]{*}{$4.835 \times 10^3$} & \multirow{2}[2]{*}{---} & \multirow{2}[2]{*}{0.623} \\
    + SVM &       &       &  \\
    \midrule
    VLAD + & \multirow{2}[2]{*}{46.21} & \multirow{2}[2]{*}{8.6} & \multirow{2}[2]{*}{0.039} \\
    Metric Learning &       &       &  \\
    \midrule
    Six Layers  & \multirow{2}[2]{*}{41.33} & \multirow{2}[2]{*}{31} & \multirow{2}[2]{*}{0.042} \\
    RNN   &       &       &  \\
    \midrule
    Spatial-temporal  & \multirow{2}[2]{*}{27.79} & \multirow{2}[2]{*}{157} & \multirow{2}[2]{*}{1.569} \\
    RNN   &       &       &  \\
    \midrule
    Proposed & \textbf{41.33} & \textbf{12} & \textbf{0.033} \\
    \bottomrule
    \end{tabular}%
\end{table}%

\subsection{Computation Time}
\label{subsec_time}
The computation time of action recognition usually includes feature extraction time, training time and testing time. We make a comparison of the computation time of selected methods, and the results are shown in Table \ref{tab:time} (the unit is second). From Table 4, we can see that the proposed method consumes the shorter total time than the other methods. It is notice that although the average training time for VLAD is shorter, the time for each epoch is not equal because the dimension of feature decreases with epoch increases.

\section{Conclusion}
\label{sec_conclude}
In this paper, we proposed a novel object activity scene description, construction and recognition method. Aiming at the limitations of existing approaches in recognize activity scenes, we proposed to partition scene into several primitive actions based upon motion attention mechanism, and then well-defined features containing both spatial and temporal information are extracted to describe these primitive actions. By regarding actions of joint as ``words'' and the corresponding activity scene as a ``paragraph'', a convolution neural network has been employed in recognition motivated by text classification. Experimental results reveal that each step of the proposed method contributes significantly to improving the accuracy of scene recognition. Through comparison with existing algorithms, we find that the propose method outperforms them in recognition accuracy and time complexity.

{
\footnotesize
\bibliographystyle{ieee}
\bibliography{jsr_v1_arxiv_release}
}

\end{document}